\documentclass[10pt,twocolumn,letterpaper]{article}

\usepackage[pagenumbers]{cvpr} 

\definecolor{cvprblue}{rgb}{0.21,0.49,0.74}
\definecolor{lavender}{RGB}{245,240,255}
\usepackage[pagebackref,breaklinks,colorlinks,allcolors=cvprblue]{hyperref}

\usepackage{multirow}
\usepackage{amsmath} 
\usepackage{booktabs} 
\usepackage{graphicx}
\usepackage{bm}
\usepackage{array}
\usepackage{colortbl}
\usepackage{xcolor}
\usepackage{titletoc}
\usepackage{makecell}
\usepackage{lineno}

\def\paperID{430} 
\def\confName{CVPR}
\def\confYear{2026}

\title{SocialNav: Training Human-Inspired Foundation Model \\ for Socially-Aware Embodied Navigation}

\author{
   Ziyi Chen$^{1}$\footnotemark[1] \and
   Yingnan Guo$^{1,2}$\footnotemark[1] \and
   Zedong Chu$^{1}$\footnotemark[2] \and
   Minghua Luo$^{1}$ \and 
   Yanfen Shen$^{1}$ \and 
   Mingchao Sun$^{1}$ \and 
   Junjun Hu$^{1}$ \and 
   Shichao Xie$^{1}$ \and 
   Kuan Yang$^{1}$ \and 
   Pei Shi$^{1}$ \and 
   Zhining Gu$^{1}$ \and 
   Lu Liu$^{1}$ \and 
   Honglin Han$^{1}$ \and 
   Xiaolong Wu$^{1}$ \and 
   Mu Xu$^{1}$ \and 
   Yu Zhang$^{2}$ \and 
   Ning Guo$^{1}$ \and 
   $^{1}$Amap, Alibaba Group, China \\
   $^{2}$Zhejiang University, China
}

\begin{document}
\maketitle

{
  \renewcommand{\thefootnote}%
    {\fnsymbol{footnote}}
  \footnotetext[1]{Equal contribution.}
  \footnotetext[2]{Corresponding author{\tt (chuzedong.czd@alibaba-inc.com)}}
}

\begin{abstract}
Embodied navigation that adheres to social norms remains an open research challenge. Our \textbf{SocialNav} is a foundational model for socially-aware navigation with a hierarchical "brain-action" architecture, capable of understanding high-level social norms and generating low-level, socially compliant trajectories. To enable such dual capabilities, we construct the SocNav Dataset, a large-scale collection of 7 million samples, comprising (1) a {Cognitive Activation Dataset} providing social reasoning signals such as chain-of-thought explanations and social traversability prediction, and (2) an Expert Trajectories Pyramid aggregating diverse navigation demonstrations from internet videos, simulated environments, and real-world robots. A multi-stage training pipeline is proposed to gradually inject and refine navigation intelligence: we first inject general navigation skills and social norms understanding into the model via imitation learning, and then refine such skills through a deliberately designed \underline{S}ocially-\underline{A}ware \underline{F}low \underline{E}xploration GRPO (SAFE-GRPO), the first flow-based reinforcement learning framework for embodied navigation that explicitly rewards socially compliant behaviors. 
SocialNav achieves \textbf{+38\%} success rate and \textbf{+46\%} social compliance rate compared to the state-of-the-art method, demonstrating strong gains in both navigation performance and social compliance. 
Our project page: {\tt\small \url{https://amap-eai.github.io/SocialNav/}}
\end{abstract}
\vspace{-1mm}

\section{Introduction}
\label{sec:intro}

\begin{figure}
    \centering
    \includegraphics[width=\linewidth]{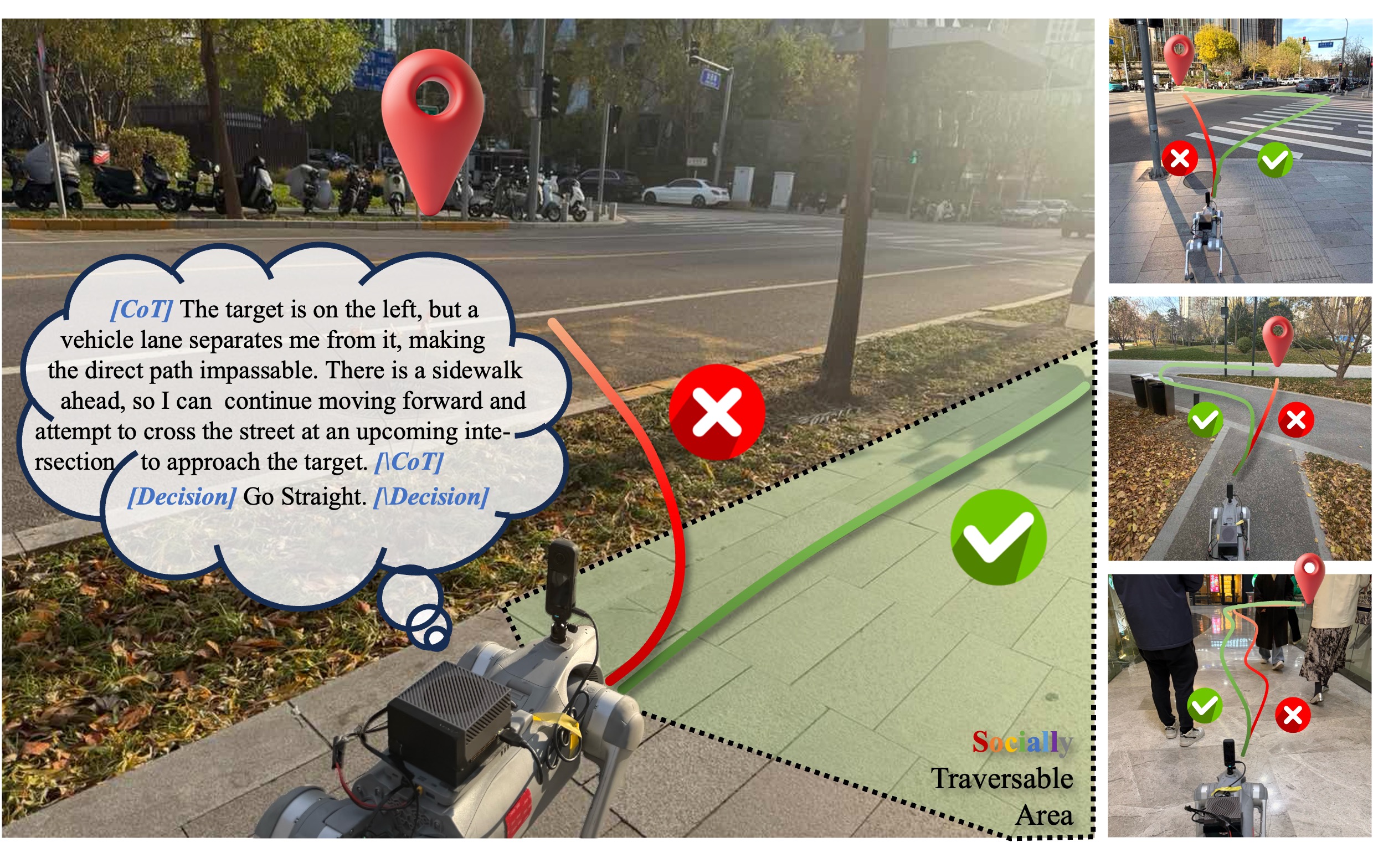} 
    \caption{\textbf{
    Socially-Aware Navigation in Real-World Environments.} SocialNav combines high-level semantic reasoning with low-level trajectory generation. It identifies socially traversable zones and generates CoT explanations, planning routes that respect social norms.
    }
    \label{fig:intro}
\end{figure}
As embodied agents become increasingly integrated into everyday social environments, robotic navigation must prioritize not only operational efficiency but also social awareness to ensure safety and compliance with established social norms. However, most existing approaches~\cite{singamaneni2024survey,shah2023gnm,shah2023vint,sridhar2024nomad,liu2025citywalker} focus primarily on shortest-path planning and collision avoidance, while overlooking the social compliance essential to real-world deployment (e.g., a robotic guide dog). Consequently, trajectories that appear optimal from a geometric or efficiency perspective may still lead to socially disruptive or inappropriate behaviors, such as jaywalking and traversing restricted zones (e.g., landscaped lawns).

To bridge this gap, we propose \textbf{SocialNav}, a hierarchical foundation model for socially aware navigation that integrates the understanding of social norms with the generation of socially compliant trajectories. {SocialNav} consists of two core components: (1) the Brain Module, built upon a vision-language model (VLM), which encodes rich social navigation priors and is capable of generating interpretable chain-of-thought (CoT) explanations or explicitly predicting socially traversable regions; and (2) the Action Expert, based on conditional flow matching~\cite{lipman2022flow}, which translates the semantic priors provided by the Brain Module into robot-executable trajectories that adhere to social norms.

Developing such a model requires rich, multimodal data that encodes both cognitive knowledge and action-oriented intuition that are currently scarce in existing embodied navigation corpora. To this end, we construct the SocNav Dataset, a large-scale heterogeneous corpus of 7 million samples that integrates two complementary modalities: (1) the Cognitive Activation Dataset (CAD), which encapsulates navigational knowledge through chain-of-thought (CoT) reasoning, social traversability prediction, and embodied question answering; and (2) the Expert Trajectories Pyramid (ETP), which aggregates trajectories from internet videos, simulated environments, and real-world robot deployments to distill rich, context-aware action priors tailored for complex social navigation scenarios.

As is widely recognized, aligning agent behavior with social norms goes beyond the representational and reasoning capabilities of standard imitation learning. Even when social priors are implicitly embedded in demonstration data, behavior cloning fails to capture the causal structure underlying normative conduct. To address this limitation, we propose Socially-Aware Flow Exploration GRPO (SAFE-GRPO), the first flow-based reinforcement learning framework for embodied navigation that explicitly promotes socially compliant behavior through norm-aware reward mechanisms. This approach enables agents to internalize the underlying principles governing social conventions, rather than merely mimicking surface-level actions.

Finally, we introduce the SocNav Benchmark, a high-fidelity evaluation platform blending physics simulation (Isaac Sim~\cite{NVIDIA_Isaac_Sim}) and photorealistic rendering (3DGS~\cite{kerbl20233d}) in 9 newly captured large-scale social scenes. Our benchmark enables comprehensive comparisons of both fundamental navigation capabilities and social compliance. Our contributions are summarized as follows:
\begin{itemize}
\item \textbf{SocialNav Foundation Model}: a hierarchical brain–action architecture that unifies high-level social norm understanding with the generation of norm-compliant trajectories.
\item \textbf{SAFE-GRPO RL Framework}: the first flow-based RL framework for embodied navigation designed to enforce social compliance through norm-aware reward shaping.
\item \textbf{SocNav Dataset \& Benchmark}: an integrated system featuring a large-scale cognition–action dataset and a high-fidelity evaluation platform built on Isaac Sim and 3DGS, providing comprehensive support for training and evaluating socially intelligent navigation models.
\end{itemize}

Extensive experiments on different benchmarks show that \textbf{SocialNav} significantly outperforms state-of-the-art baselines in open-loop, closed-loop, and real-world evaluation. Crucially, it demonstrates superior social compliance, taking a meaningful stride towards true embodied social intelligence.

\vspace{-1mm}

\section{Related Work}
\label{sec:related_work}

\subsection{Visual Navigation}
Visual navigation~\cite{bonin2008visual} has evolved from classical SLAM-based methods~\cite{cadena2017past, taketomi2017visual, kazerouni2022survey, tsardoulias2016review} to end-to-end learning approaches such as GNM~\cite{shah2023gnm}, ViNT~\cite{shah2023vint}, and NoMaD~\cite{sridhar2024nomad}. To improve generalization, a line of work has expanded training corpora beyond expert demonstrations. Some methods, such as CityWalker~\cite{liu2025citywalker} and MBRA~\cite{hirose2025mbra}, leverage massive internet-sourced video collections to capture diverse action priors, while others~\cite{he2025seeingexperiencingscalingnavigation, cheng2024navila} employ simulation platforms~\cite{wu2025urbansim, savva2019habitat} to generate controlled, rule-compliant trajectories at scale. Concurrently, VLMs are used to enhance semantic understanding~\cite{rajvanshi2024saynav,cheng2025navila,zhang2024navid, wang2025trackvla,xue2025omninav}. However, these directions still struggle with high-level reasoning and alignment with human values in socially complex scenarios.

\subsection{Social Navigation and Human Preference Alignment}
Social navigation requires adherence to social conventions, moving beyond early handcrafted costs~\cite{Chen2017,roth2024viplanner} to VLM-based reasoning~\cite{lin2025navcot,wang2025think, long2024discuss}. However, VLM reasoning is often disconnected from low-level action generation. Recently, flow matching (FM)~\cite{lipman2022flow} has been widely adopted in vision-and-language-action (VLA) models~\cite{intelligence2025pi05visionlanguageactionmodelopenworld,yang2025nav} for its ability to model multimodal action distributions, but these models are typically limited to behavior cloning. In the navigation domain, pure imitation learning often lacks the causal understanding needed to robustly adapt to novel or dynamic social situations. Therefore, there is an inherent need for models that can not only generate actions but also align with complex human preferences and social norms. Foundational work, such as GRPO~\cite{guo2025deepseek, Shao2024} and Flow-GRPO~\cite{liu2505flow}, demonstrates combining generative models with online RL for human preference alignment, inspiring our approach. 

\section{SocNav Dataset and Benchmark}\label{sec:data_benchmark}

\begin{figure*}[t]
    \centering
    \includegraphics[width=\textwidth]{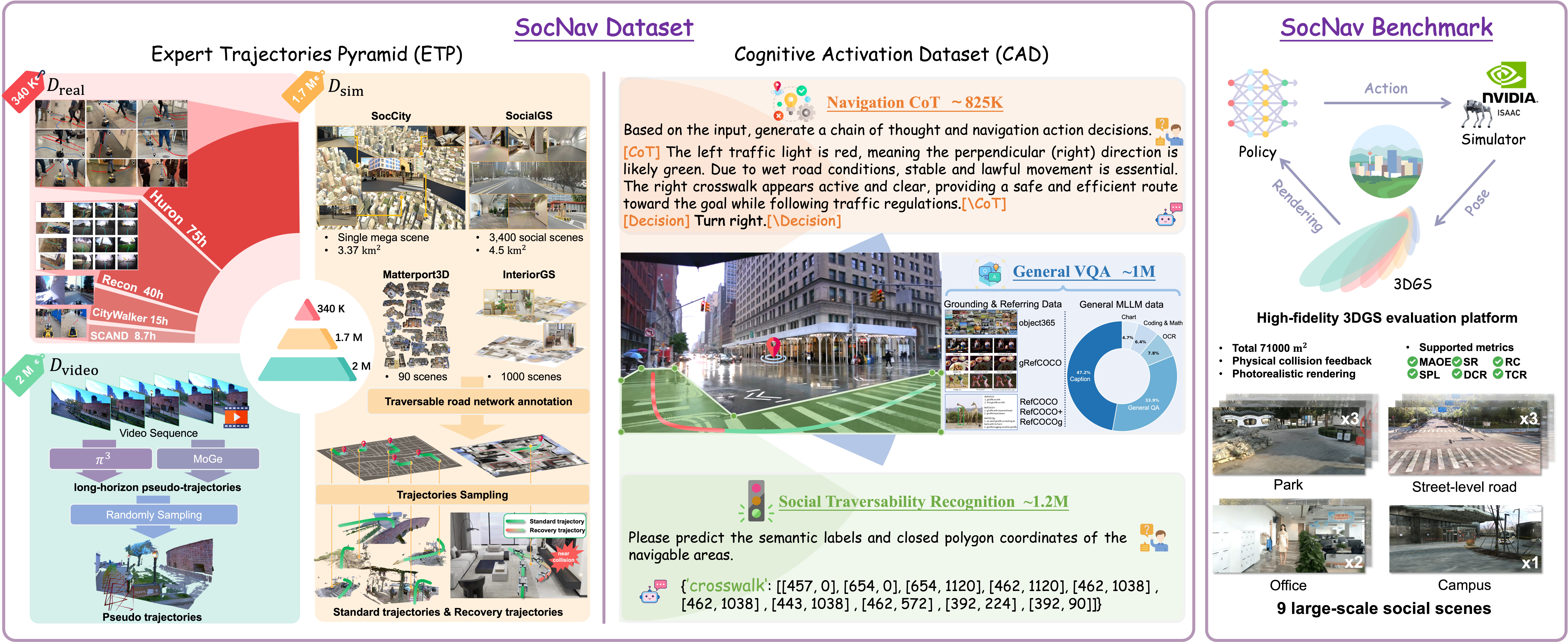}      
     \caption{\textbf{Overview of the SocNav Dataset and Benchmark.} The SocNav Dataset (left) illustrates the hierarchical structure and data construction pipeline, composed of the Expert Trajectories Pyramid (ETP) and Cognitive Activation Dataset (CAD). The SocNav Benchmark (right) is a high-fidelity evaluation platform featuring large-scale, diverse social environments and offering comprehensive metrics to assess socially-aware navigation performance.}
    \label{fig:dataset_overview}
\end{figure*}

To enable the development and evaluation of socially-aware embodied navigation systems, we introduce the \textbf{SocNav Dataset} and the \textbf{SocNav Bench}. 
Together, they form a comprehensive ecosystem for training, validating, and benchmarking embodied agents.

\subsection{SocNav Dataset}
\label{sec:data}
We present the \textbf{SocNav Dataset}, a large-scale, multi-source heterogeneous dataset to support robust, generalizable, and socially-aware navigation. 
Comprising over \textbf{7 million} training samples, the SocNav Dataset consists of two core components: the {Expert Trajectories Pyramid (ETP)} for learning motion priors and the {Cognitive Activation Dataset (CAD)} for developing high-level reasoning.

\noindent \textbf{Expert Trajectories Pyramid (ETP).}\quad
The ETP is organized as a three-layer hierarchical data pyramid, covering trajectories from diverse indoor spaces (e.g., homes, offices, malls) and outdoor urban scenes (e.g., streets, parks, open areas).  We define a training trajectory sample as a point-goal navigation instance.
As illustrated in Figure~\ref{fig:dataset_overview}, ETP consists of:

\begin{itemize}
\item\noindent \textbf{Trajectories from internet video ($D_\text{video}$).} 
$D_\text{video}$, the foundational layer, comprises \textbf{2.0 million} pseudo-trajectories, meticulously curated from extensive public urban exploration videos to ensure vast visual diversity across global cities, weather conditions, and architectural styles. These raw video streams are transformed into pseudo-trajectories via a scalable pipeline, involving: (1) Dense 3D reconstruction using $\pi^{3}$~\cite{wang2025pi}, (2) Metric scale alignment through MoGe~\cite{wang2025moge}, and (3) Pseudo-trajectory synthesis by sampling diverse point-goal episodes along the reconstructed paths.

\item \noindent \textbf{Trajectories from High-Fidelity Simulated Scenes ($D_\text{sim}$).}\quad
This middle data layer provides 1.7 million diverse trajectories. It comprises over \textbf{4,490} high-fidelity 3D scenes and a large-scale dynamic city. Our primary contributions to this layer are: (1) \textbf{SocialGS}, a new dataset of \textbf{3,400} real-world scenes (4.5 km²) reconstructed via 3DGS. It diversifies beyond existing residential datasets (\textit{e.g.}, Matterport3D~\cite{ramakrishnan2021habitat}, InteriorGS~\cite{InteriorGS2025}) by covering social environments like shopping malls, streets, and offices. (2) \textbf{SocCity}, a \textbf{3.37 km²} dynamic urban scene with simulated vehicles and pedestrians in Isaac Sim. 
Across all scenes, trajectories are generated on manually annotated traversable road networks. These trajectories encompass not only standard on-road navigation but also challenging recovery scenarios, such as near-collisions. This rich dataset enables the model to learn both efficient navigation and robust recovery behaviors.

\item\noindent \textbf{Trajectories from real-world robot data ($D_\text{real}$).} This layer provides \textbf{340K} high-quality trajectories collected from autonomous robots deployed in real-world environments (from public datasets including SCAND~\cite{karnan2022socially}, Huron~\cite{hirose2023sacson}, Recon~\cite{shah2022rapid}, and CityWalker teleoperation data~\cite{liu2025citywalker}). These trajectories provide ground-truth metric accuracy, physical realism, and sensor consistency, capturing true physical dynamics, sensor noise, and environmental interactions. They are ideal for supervised fine-tuning (SFT) and closing the sim-to-real gap.
\end{itemize}

\begin{figure*}[t]
    \centering
    \includegraphics[width=\textwidth]{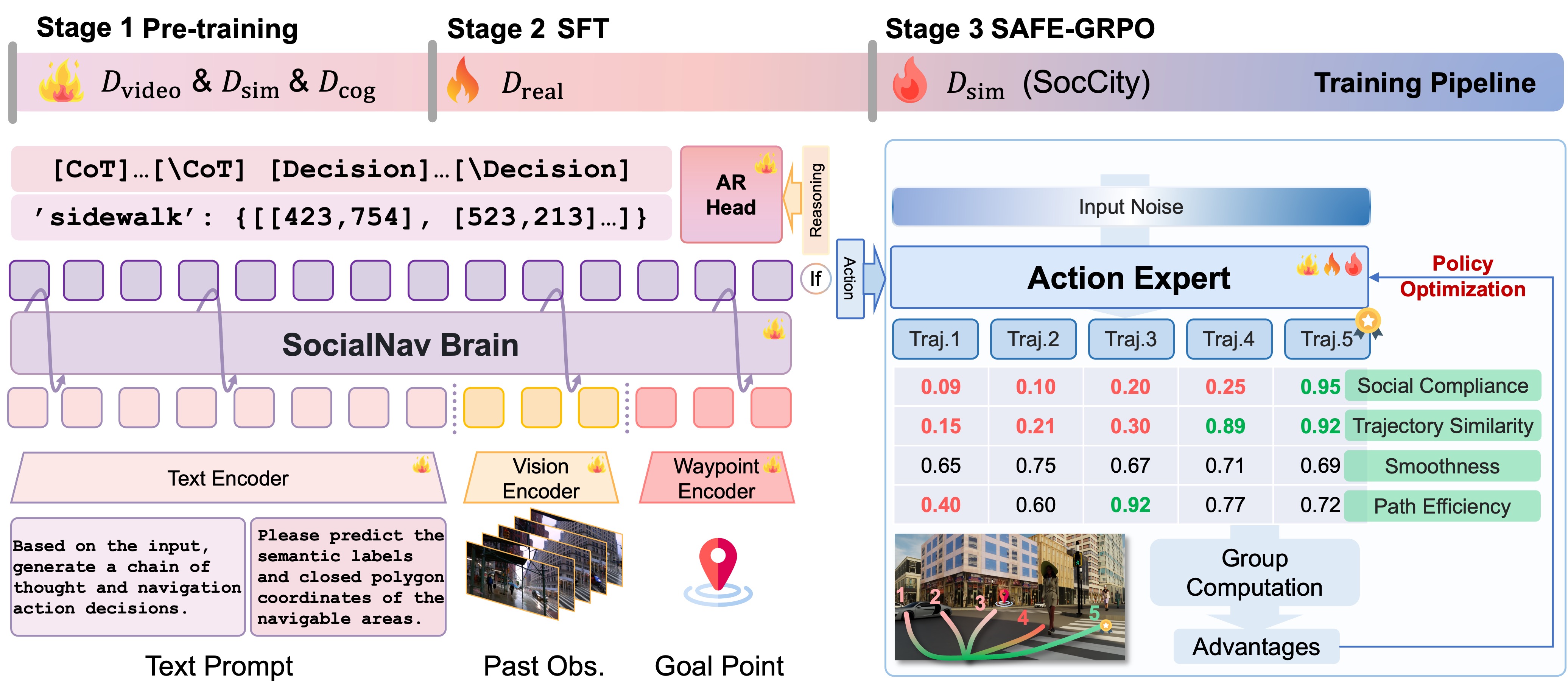}      
    \caption{\textbf{SocialNav Architecture and Training Pipeline.} SocialNav adopts a hierarchical architecture, with a VLM-based Brain for high-level semantic reasoning and an action expert for generating socially compliant trajectories. We adopt a three-stage training strategy: Pre-training, Fine-tuning, and SAFE-GRPO.}
    \label{fig:model}
\end{figure*}

\noindent \textbf{Cognitive Activation Dataset (CAD).}\quad
To cultivate advanced spatial cognition beyond geometric path planning, we introduce the Cognitive Activation Dataset ($D_\text{cog}$). It integrates a suite of non-trajectory auxiliary tasks designed to instill a semantically grounded understanding of environmental rules and human-centric norms.
\begin{itemize}
\item \textbf{Social Traversability Recognition:} We created a dataset of \textbf{1.2 million} samples by manually annotating polygons over socially traversable regions (e.g., sidewalks, crosswalks, park trails) in first-person images from internet video. This directly trains the model to distinguish between physically and socially acceptable paths.
\item \textbf{Navigation Chain-of-Thought (CoT):} Leveraging first-person images and curated prompt templates, we prompted Qwen2.5VL-72B~\cite{bai2025qwen2} to generate 825K Chain-of-Thought (CoT) samples. These samples, comprising step-by-step textual rationales for navigation decisions, were designed to teach the agent explicit reasoning.
\item \textbf{General Visual Question Answering (VQA):} To ensure the model maintains general-world knowledge, we curated \textbf{1 million} general VQA samples from \cite{guo2025mammoth,chen2025revisiting,chen2025blip3o,lin2014microsoft,liu2023gres,shao2019objects365}. This task trains the agent to reason about spatial relationships and object properties in the environment.
\end{itemize}

Together, the ETP and CAD establish the \textbf{SocNav Dataset} as a comprehensive foundation for training foundational navigation agents—unifying scale, realism, and cognition within a single, coherent data framework.

\subsection{SocNav Benchmark}\label{sec:benchmark}
We introduce the \textbf{SocNav Bench}, a unified, high-fidelity evaluation platform for socially-aware navigation. It achieves a unique blend of realism by combining the physics simulation of Isaac Sim with the photorealistic rendering of 3DGS. The benchmark is built upon 9 new large-scale social scenes we captured and reconstructed using 3DGS, covering a total area of \textbf{73K $m^2$}. These new scenes include diverse, human-centric environments: three parks, three street-level roads, two offices, and one campus. All evaluations employ a standardized setup: a unified Unitree Go2 robot model and a consistent locomotion policy. To enable realistic physical interaction, each 3DGS scene is converted into a mesh and imported into Isaac Sim for accurate collision feedback. Additionally, to simulate potential dynamic collisions, digital humans are randomly introduced into the scenes. For rigorous benchmarking, we sample 10 start-goal pairs at distances of 20m and 100m within each scene, creating 20 evaluation cases per scene.

\section{SocialNav Foundation Model}
\label{sec:main_sec}

\subsection{Problem Formulation}
We formulate the foundational navigation task as a vision-based, history-conditioned point-goal navigation problem in diverse environments, similar to the setup in CityWalker~\cite{liu2025citywalker}. At each time step $t$, the agent receives a sequence of recent monocular visual observations
$\bm{O}_{t-n:t}=\{\bm{o}_{t-n},\ldots,\bm{o}_t\}$, where $\bm{o} \in \mathbb{R}^{H \times W \times 3}$, along with their associated 2D positional information $\bm{P}_{t-n:t}=\{\bm{p}_{t-n},\ldots,\bm{p}_t\}$ with $\bm{p} \in \mathbb{R}^{2}$. Given a specified 2D target location $\bm{g} \in \mathbb{R}^{2}$, the objective is to learn a policy $\pi_{\theta}$ that maps the historical observations and positions to a sequence of future actions:
\begin{equation}
    \bm{A}_{t+1:t+m} = \pi_{\theta}(\bm{O}_{t-n:t},\, \bm{P}_{t-n:t},\, \bm{g}),
    \label{eq:definition}
\end{equation}
Unless otherwise specified, we set $n=5$ and $m=5$.

\subsection{Model Overview}
Our complete model architecture, illustrated in \cref{fig:model}, adopts a hierarchical "brain-action" design for robust and socially aware embodied navigation. It consists of two tightly coupled branches with complementary roles: one for high-level semantic understanding and the other for low-level action generation.

\vspace{+1.5mm}
\noindent \textbf{The Brain Module.}\quad
The Brain Module serves as the cognitive core of the system, implemented as a Vision–Language Model (VLM) denoted by $\pi_\text{VLM}$. It performs generative, autoregressive textual reasoning to infer critical environmental semantics and can produce 3 types of interpretable outputs:
\begin{itemize}
\item \textbf{Socially traversable regions:} Represented as polygons, these delineate areas such as sidewalks, crosswalks, and stairs.
\item \textbf{CoT reasoning:} Step-by-step textual explanations for navigational decisions.
\item \textbf{VQA:} Responses to free-form questions that enhance scene understanding.
\end{itemize}
By explicitly generating textual outputs, the VLM provides interpretable insights essential for safe and socially responsible navigation. This design enables the agent not only to \textit{see} but also to \textit{reason about} its surroundings, forming the cognitive foundation for socially aware navigation.

\noindent \textbf{The Action Expert.}\quad
This action expert specializes in end-to-end trajectory generation. Inspired by recent advancements~\cite{black2410pi0, intelligence2025pi05visionlanguageactionmodelopenworld} in action prediction, we leverage conditional flow matching~\cite{lipman2022flow} to model a distribution of actions. The module is conditioned on latent semantic features extracted from the last-layer features of the VLM:
\begin{gather}
    \bm{A}_{t+1:t+m} = \pi_\text{flow}(\bm{x}_t, t; \bm{Z}_{\text{VLM}}) \\
    \bm{Z}_{\text{VLM}} = \pi_\text{VLM}(\bm{O}_{t-n:t}, \bm{P}_{t-n:t}, \bm{g})
\end{gather}
This mechanism enables the action expert to produce efficient and socially compliant trajectories in complex environments by decoupling high-level reasoning from low-level control while preserving a strong semantic connection between them.

\begin{table*}[t]
    \centering
    \caption{\textbf{Open-Loop Evaluation on CityWalker Benchmark}~\cite{liu2025citywalker}. We evaluate MAOE metric in each critical scenario for all methods. Percentages under scenarios indicate their data proportions. The "Mean" column shows scenario means averaged over six scenarios; "All" shows sample means over all data samples.}\label{tab:benchmark}
    \begin{tabular}{l|cccccccc}
    \toprule

        \multirow{2}{*}{\textbf{Method}} & \multirow{2}{*}{\textbf{Mean}} & \textbf{Turn} & \textbf{Crossing} & \textbf{Detour} & \textbf{Proximity} & \textbf{Crowd} & \textbf{Other} & \textbf{All} \\ 

        & & 8\% & 12\% & 12\% & 6\% & 7\% & 55\% & 100\% \\ 
        \midrule
        GNM~\cite{shah2023gnm}
         & 16.2 & 31.1 & 14.8 & \underline{12.5} & 14.7 & 12.8 & 11.0 & 12.1\\

        ViNT~\cite{shah2023vint}
         & 16.5 & 31.1 & 15.4 & 12.9 & 14.8 & 13.3 & 11.6 & 12.6\\

        NoMaD~\cite{sridhar2024nomad}
         & 19.1 & 35.1 & 18.5 & 15.6 & 18.1 & 14.3 & 12.8 & 12.1\\

        CityWalker~\cite{liu2025citywalker} 
         &  \underline{15.2} &  \underline{26.6} &  \underline{14.1}  &  13.9 &  \underline{14.3} &  \underline{12.0} &  \underline{10.4} &  \underline{11.5}\\
        \rowcolor{lavender}
        \textbf{SocialNav (Full)}
         &  \textbf{10.2} & \textbf{20.1} &  \textbf{8.8}  &  \textbf{8.4} &  \textbf{8.9} &  \textbf{7.6} &  \textbf{7.2} &  \textbf{7.8}\\ 
    \bottomrule
    \end{tabular}
\end{table*}

\begin{table*}[htbp]
\centering
\caption{\textbf{Performance Comparison on the Closed-Loop SocNav Benchmark}.}\label{tab:gs_performance}

{

\newcolumntype{C}{>{\centering\arraybackslash}p{1.5cm}}

\begin{tabular}{l|CCC|CC} 
    \toprule
    & \multicolumn{3}{c|}{\textbf{Navigation Performance}} & \multicolumn{2}{c}{\textbf{Social Compliance}} \\
    \cmidrule(lr){2-4} \cmidrule(lr){5-6} 
    \textbf{Methods} & \textbf{SR$\uparrow$} & \textbf{RC$\uparrow$} & \textbf{SPL$\uparrow$} & \textbf{DCR$\uparrow$} & \textbf{TCR$\uparrow$}  \\
    \midrule
    GNM*~\cite{shah2023gnm} & 43.3 & 62.4 & 37.0 & 26.5 & 28.7 \\
    ViNT*~\cite{shah2023vint} & 45.6 & \underline{66.2} & 39.5 & 31.4 & 33.8   \\
    NoMaD*~\cite{sridhar2024nomad} & 41.1 & 60.5 & 35.4 & 29.5 & 31.6  \\
    CityWalker~\cite{liu2025citywalker} & \underline{47.8} & 64.7 & \underline{44.7} & \underline{36.1} & \underline{36.6} \\
    SocialNav* & 65.0 & 78.4 & 62.3 & 58.0 & 56.7 \\
   \rowcolor{lavender}
   \textbf{SocialNav (Full)} & \textbf{86.1}  & \textbf{91.2} & \textbf{77.4} & \textbf{82.5} & \textbf{82.9}  \\
    \bottomrule
\end{tabular}
}
\end{table*}

\subsection{Training Methodology}
Our training pipeline follows a multi-stage strategy designed to progressively instill both general navigation priors and social compliance into {SocialNav}. 

\noindent \textbf{Stage 1: Pre-training for General Navigation Ability.}\quad
In the first stage, we aim to activate the VLM’s navigation capability and train the flow model to predict low-level waypoints. This is achieved through pretraining on the ETP datasets ($D_\text{video}$ and $D_\text{sim}$) together with the cognitive activation dataset $D_\text{cog}$.
$D_\text{video}$ provides diverse real-world navigation scenarios with implicit expert behaviors, while $D_\text{sim}$ introduces challenging synthetic cases to enhance robustness in rare and complex situations. The $D_\text{cog}$ further improves the VLM’s reasoning and decision-making through CoT and VQA tasks, and equips it with the ability to predict traversable regions—laying a solid foundation for subsequent social-norm alignment.

\noindent \textbf{Stage2: Fine-tuning with High-Quality Real-World Data.}\quad
In the second stage, we fine-tune the model on high-quality expert trajectories collected from real-world robots ($D_\text{real}$) to reduce the sim-to-real gap. During this phase, the VLM is frozen, and only the action expert is optimized. This approach preserves the VLM’s semantic and social reasoning capabilities while allowing the flow model to adapt to real-world dynamics and spatial scales.

\noindent \textbf{Stage3: Reinforcement Learning for Social Rule Alignment.}\quad
Although the previous stages equip the model with strong navigation priors and real-world adaptability, imitation learning still lacks causal reasoning in social environments. We therefore introduce \textbf{SAFE-GRPO} (Socially-Aware Flow Exploration GRPO), a reinforcement learning stage that explicitly aligns the policy with human social conventions. The model is trained using expert trajectories from the SocCity within $D_\text{sim}$, which provides accurate and rich pathway annotations crucial for precise reward feedback. 

To encourage diverse yet meaningful exploration, we draw inspiration from the flow-based formulation of \textit{Flow-GRPO}~\cite{liu2505flow, wang2025cps}, converting the deterministic ordinary differential equation (ODE) of the flow policy into a stochastic differential equation (SDE):
\begin{equation}
d\bm{x}_t = \bm{v}_\text{flow}(\bm{x}_t, t; \bm{Z}_{\text{VLM}})dt + \sigma_td\bm{w}_t,
\label{eq:sde_flow}
\end{equation}
where $\sigma_t$ controls the exploration magnitude. Here, $\bm{v}_\text{flow}$ represents the velocity field of the flow policy, which is conditioned on both the current state $\bm{x}_t$ and time $t$, as well as additional context $\bm{Z}_{\text{VLM}}$ provided by the VLM.

Unlike unstructured stochastic exploration, our approach is controlled and semantically grounded: randomness is introduced only during flow integration, while the semantic conditioning signal derived from the VLM "Brain" remains fixed throughout. This latent prior encodes high-level spatial and social cues. Without explicit comprehension of scene semantics and walkable areas, conventional RL agents struggle to formulate or discover socially-compliant behaviors solely from inefficient exploration and sparse rewards.

Trajectories that are collision-free and socially valid receive higher rewards, reinforcing alignment between action generation and human expectations. The overall reward balances social compliance and navigational efficiency:
\begin{equation}
\mathcal{R} =
\mathcal{R}_\text{social} +
\lambda_\text{expert}\mathcal{R}_\text{expert} +
\lambda_\text{smooth}\mathcal{R}_\text{smooth} +
\lambda_\text{eff}\mathcal{R}_\text{eff},
\label{eq:safegrpo_reward}
\end{equation}
where $\mathcal{R}_{\text{social}}$ is the primary reward, derived from a semantic occupancy map $\mathcal{M}_{\text{occ}}$ that encourages the agent to maintain safe clearance from all non-traversable areas,
$\mathcal{R}_{\text{expert}}$ promotes consistency with expert trajectories, $\mathcal{R}_{\text{smooth}}$ enforces natural motion continuity, and $\mathcal{R}_{\text{eff}}$ rewards efficient progress toward the goal.
Together, these terms ensure that the agent navigates effectively while remaining socially compliant and human-aligned. Detailed formulations are provided in the Appendix.

\section{Experiment}
\label{sec:experiment}

To ensure a fair comparison, we evaluate {SocialNav} in three distinct settings: (1) open-loop benchmark proposed by CityWalker~\cite{liu2025citywalker}, (2) closed-loop evaluation on our SocNav Benchmark, and (3) real-world robotic deployments.

\subsection{Setup}

\noindent \textbf{Metrics.}\quad For open-loop evaluation, we use Maximum Average Orientation Error (MAOE) following CityWalker~\cite{liu2025citywalker}. Closed-loop performance is measured by success rate (SR; defined as reaching within 3 meters of the goal with fewer than three collisions), route completion (RC), and success weighted by path length (SPL). To assess social compliance, we introduce the Distance Compliance Rate (DCR) and Time Compliance Rate (TCR):
\begin{equation}
\text{DCR} =
\begin{cases}
    \dfrac{d_\text{compliant}}{d_\text{actual}}, & \text{if } s = 1 \\[8pt]
    0, & \text{otherwise}
\end{cases}
\end{equation}
where $s$ is a binary indicator (1 for success, 0 for failure), ${d}_{\text{compliant}}$ denotes the distance traveled within socially compliant regions, and ${d}_{\text{actual}}$ represents the total actual distance traveled. The TCR is formulated similarly.


\begin{figure*}[t]
    \centering
    \includegraphics[width=\textwidth]{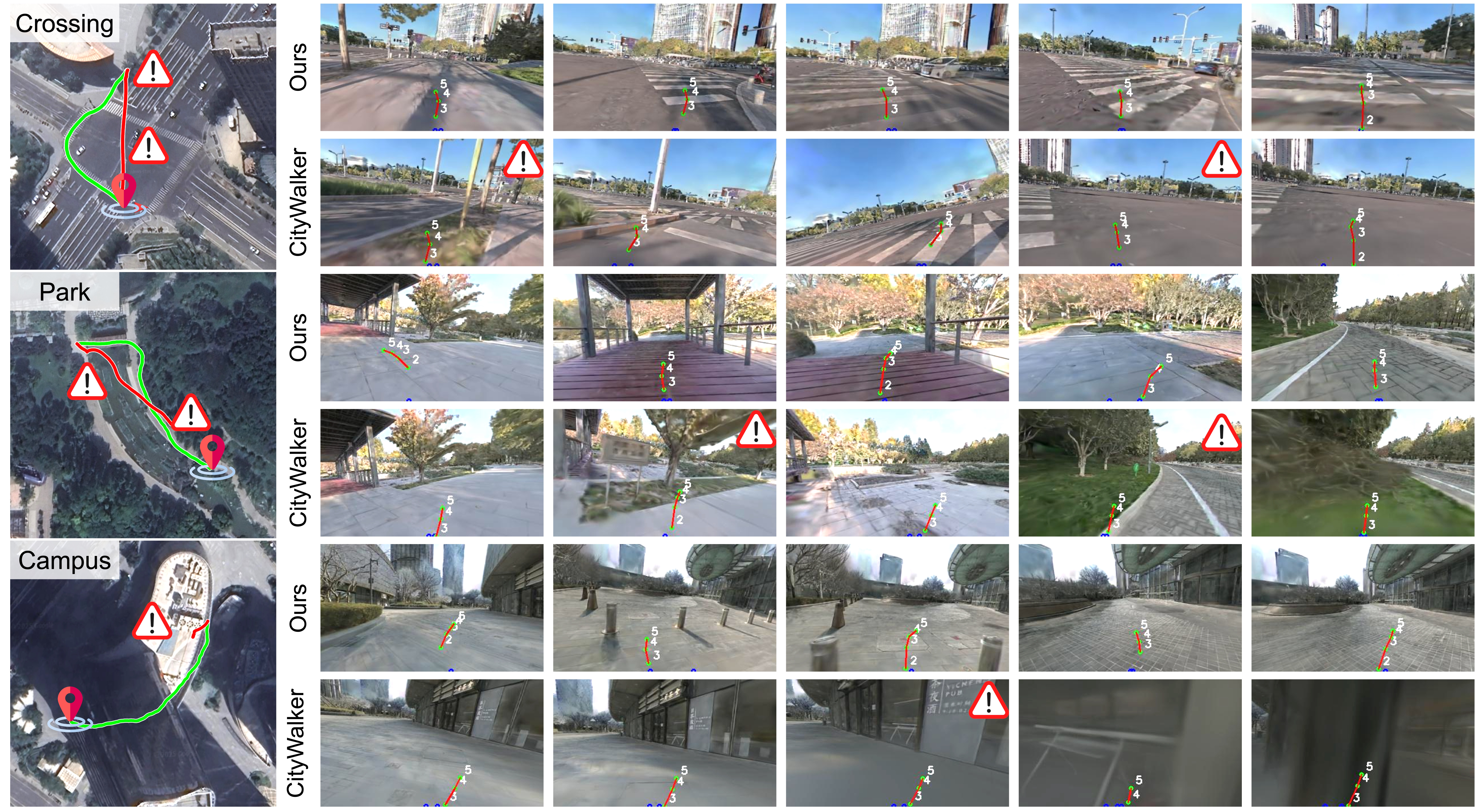} 
    \caption{\textbf{Qualitative comparison on the SocNav Benchmark.} We visualize representative trajectories in three scenes (Crossing, Park, Campus). The left column shows top-down path views with our method (green) and the CityWalker baseline (red), where warning signs mark unsafe or socially improper behaviors. The right columns depict corresponding egocentric views: SocialNav remains on sidewalks and walkways, while the baseline often takes shorter but socially risky routes through restricted regions (such as driveways, dry streambeds, lawns, and green belts) or crashes into obstacles like glass walls and trees.}
    \label{fig:qualitative_vis}
\end{figure*}

\noindent \textbf{Model Details.}\quad We employ Qwen2.5-VL (3B) as the brain module. The action expert is designed as a Diffusion Transformer~\cite{peebles2023scalable} with $L=12$ layers, $H=12$ attention heads per layer, and hidden dimension $D=1536$. During inference, the trajectory denoising is performed iteratively for $K=5$ steps.

\noindent \textbf{Training Details.}\quad Pre-training Stage: The full model is trained end-to-end using AdamW~\cite{loshchilov2017decoupled} (3 epochs, 96 H20 GPUs) with a batch size of 192 and a learning rate of $5 \times 10^{-5}$. Finetuning Stage: Only the action expert is fine-tuned on 32 H20 GPUs with a batch size of 256 and a learning rate of $1 \times 10^{-5}$.
SAFE-GRPO: We further optimize the action expert on 16 H20 GPUs, using a rollout batch size of 128 and a learning rate of $5 \times 10^{-7}$.

\subsection{Baselines}

We compare SocialNav against state-of-the-art (SOTA) open-source point-based navigation methods. The selected baselines include CityWalker~\cite{liu2025citywalker}, ViNT~\cite{shah2023vint}, GNM~\cite{shah2023gnm}, and NoMaD~\cite{sridhar2024nomad}.
While ViNT, GNM, and NoMaD were originally designed for image-goal navigation, we re-trained them for point-goal tasks to ensure a fair comparison. In \cref{tab:gs_performance} and \cref{tab:real_world_results}, an asterisk (*) denotes the models that were exclusively trained on the $D_\text{real}$ dataset.

\subsection{Results on CityWalker Benchmark}
We evaluate on the open-loop benchmark introduced by CityWalker~\cite{liu2025citywalker}. As shown in \cref{tab:benchmark}, our approach consistently outperforms prior methods across key scenarios, demonstrating improved generalization and robustness. Since ground truth trajectories are human-operated, these results indicate that our method more closely aligns with human social walking norms.

\subsection{Results on our SocNav Benchmark}
We conduct closed-loop evaluation on our SocNav Benchmark, with all environments \textbf{unseen during training}. Quantitative results are shown in \cref{tab:gs_performance} and qualitative visualizations in \cref{fig:qualitative_vis}.

\noindent \textbf{Navigation Performance.} \quad
SocialNav achieves state-of-the-art results across all navigation metrics. Specifically, it significantly outperforms CityWalker~\cite{liu2025citywalker}, the second-best method, with the concurrent increase of +{38.3} in SR, +{26.5} in RC and +{32.7} in SPL.

\noindent \textbf{Social Compliance.} \quad
SocialNav achieves remarkable improvements in social compliance, attaining a DCR of {82.5} and a TCR of {82.9}, both more than double those of CityWalker (DCR: {36.1}, TCR: {36.6}), and significantly better than other baselines. Representative qualitative visualizations in \cref{fig:qualitative_vis} further highlight these advantages: SocialNav consistently selects paths that adhere to sidewalks and designated walkways, whereas the baseline frequently opts for shorter but socially inappropriate routes traversing restricted areas. This substantial improvement validates our model’s ability to internalize complex social norms. Importantly, these gains in social compliance are attained without sacrificing navigation performance.

\noindent\textbf{Effect of Our Model design.} \quad
The SocialNav* and baselines (including \text{NoMaD}*, \text{GNM}*, \text{ViNT}*) are trained with Imitation Learning (IL) solely on $D_\text{real}$. As shown in \cref{tab:gs_performance}, SocialNav* significantly outperforms the baselines, clearly demonstrating the effectiveness and superior generalization capability of our model architecture.

\begin{table}[htbp]
\centering
\caption{\textbf{Real-world Results.} Comparison of success rates across different real-world environments.}
\label{tab:real_world_results}
\resizebox{1.0\linewidth}{!}{
\begin{tabular}{l|ccc|c}
\toprule
\multirow{2}{*}{\textbf{Method}} 
& \textbf{Street} & \textbf{Office} & \textbf{Shopping} 
& \multirow{2}{*}{\textbf{Average SR}} \\
& \textbf{Crossing} & \textbf{Park} & \textbf{Mall} &  \\
\midrule
GNM*~\cite{shah2023gnm} & 9/20 & 10/20 & 8/20 & 45.0 \\
ViNT*~\cite{shah2023vint} & 7/20 & 12/20 & 8/20 & 45.0 \\
NoMaD*~\cite{sridhar2024nomad} & 9/20 & 11/20 & 10/20 & 50.0 \\
CityWalker~\cite{liu2025citywalker} & 12/20 & 13/20 & 12/20 & 62.5 \\
\midrule
\rowcolor{lavender}
\textbf{SocialNav (Full)} & \textbf{18/20} & \textbf{16/20} & \textbf{17/20} & \textbf{85.0} \\
\bottomrule
\end{tabular}
}
\end{table}

\begin{table*}[htbp]
\centering
\caption{\textbf{Ablation Study on Data Composition and Training Stages.} 
IL: Imitation Learning (Stages 1–2); RL: SAFE-GRPO (Stage 3). SocialNav* is trained with IL only on $D_\text{real}$.} 
\label{tab:ablation_data_training}
\resizebox{\textwidth}{!}{
\begin{tabular}{c|l|cccc|cc|ccc|cc}
    \toprule
    \textbf{No.} & \textbf{Model} 
        & \textbf{$D_\text{real}$} 
        & \textbf{$D_\text{video}$} 
        & \textbf{$D_\text{sim}$} 
        & \textbf{$D_\text{cog}$} 
        & \textbf{IL} 
        & \textbf{RL} 
        & \textbf{SR$\uparrow$} 
        & \textbf{RC$\uparrow$} 
        & \textbf{SPL$\uparrow$} 
        & \textbf{DCR$\uparrow$} 
        & \textbf{TCR$\uparrow$} \\
    \midrule
    1 & SocialNav*        
        & $\checkmark$ &           &            &            & $\checkmark$ &            
        & 65.0 & 78.4 & 62.3 & 58.0 & 56.7 \\ 
    2 & + $D_\text{video}$          
        & $\checkmark$ & $\checkmark$ &            &            & $\checkmark$ &            
        & 76.7 & 84.8 & 70.1 & 62.9 & 64.6 \\
    3 & + $D_\text{sim}$                         
        & $\checkmark$ & $\checkmark$ & $\checkmark$ &            & $\checkmark$ &            
        & 82.2 & 86.0 & \underline{77.8} & 69.8 & 68.2 \\
    4 & + $D_\text{cog}$                         
        & $\checkmark$ & $\checkmark$ & $\checkmark$ & $\checkmark$ & $\checkmark$ &            
        & \underline{84.4} & 88.1 & \textbf{79.4} & \underline{78.2} & \underline{78.4} \\
    5 & + RL (no $D_\text{cog}$)        
        & $\checkmark$ & $\checkmark$ & $\checkmark$ &            & $\checkmark$ & $\checkmark$
        & 80.0 & \underline{89.1} & 78.9 & 68.1  & 66.9 \\
        \rowcolor{lavender}
    6 & + RL (full data)               
        & $\checkmark$ & $\checkmark$ & $\checkmark$ & $\checkmark$ & $\checkmark$ & $\checkmark$
        & \textbf{86.1}  & \textbf{91.2} & {77.4} & \textbf{82.5} & \textbf{82.9} \\
    \bottomrule
\end{tabular}
}
\end{table*}

\subsection{Results on the Real-World Deployments}
We deploy different methods on a cloud server with an NVIDIA A10 GPU to guide a Unitree Go2 robot's navigation. We evaluate their performance on a set of point-goal navigation tasks across three distinct environments: (1) a street crossing, (2) an office park, and (3) a shopping mall. To ensure a fair comparison, we conduct 20 trials in each environment, using the same starting and ending points for all methods.
As shown in \cref{tab:real_world_results}, SocialNav achieves a SR of {85}, significantly outperforming both NoMaD~\cite{sridhar2024nomad} (50) and CityWalker~\cite{liu2025citywalker} (62.5). Notably, SocialNav attains {18/20} successful trials in the street crossing environment, reflecting robust behavior in challenging, socially constrained settings. Additionally, SocialNav operates at over 5\,Hz during deployment, enabling real-time navigation.

\subsection{Ablation Studies}

\noindent\textbf{Effect of Data Composition.}\quad
To dissect the impact of various data components on SocialNav's performance, we conduct an ablation study by progressively adding $D_\text{video}$, $D_\text{sim}$, and $D_\text{cog}$ into the IL pipeline(No.~1–4 in \cref{tab:ablation_data_training}).

\begin{itemize}
\item \textbf{Effect of $D_\text{video}$ (No.2 vs. No.1)}: Large-scale Internet trajectories provide diverse human motion patterns, yielding clear gains in navigation performance (SR +11.7, RC +6.4, SPL +7.8) and improvements in social compliance (DCR +4.9, TCR +7.9).
\item \textbf{Effect of $D_\text{sim}$ (No.3 vs. No.2)}: Expert recovery trajectories from simulation considerably strengthen robustness, leading to further improvements in SR (+5.5), RC (+1.2) and SPL (+7.7), along with moderate gains in social compliance.
\item \textbf{Effect of $D_\text{cog}$ (No.4 vs. No.3)}: Incorporating $D_\text{cog}$ yields a significant increase in social compliance rates (DCR +8.4, TCR +10.2), along with gains in overall navigation metrics. This confirms that social cognitive knowledge, by enabling CoT reasoning and traversable area prediction, is critical for the model to internalize social rules.
\end{itemize}

\noindent\textbf{Effect of RL training.} 
\begin{itemize}
\item \textbf{Effect of SAFE-GRPO (No.6 vs. No.4)}: Building on IL, SAFE-GRPO significantly improves social compliance, improving DCR from 78.2 to {82.5} and TCR from 78.4 to {82.9}. It also enhances navigation performance, achieving the highest SR {86.1} and RC 91.2. This demonstrates that flow-based RL with norm-aware rewards is effective for refining socially acceptable behaviors.
\item \textbf{The Crucial Role of $D_\text{cog}$ in RL (No.5 vs. No.3 \& No.6)}: Applying SAFE-GRPO without cognitive priors (No.5) worsens social metrics than IL-only training (with DCR -1.7 and TCR -1.3). This reveals a crucial insight: without the high-level social understanding encoded in $D_{\text{cog}}$, the RL agent lacks the foundational "brain" knowledge necessary to properly align its policy optimization with human social norms. Conversely, combining $D_{\text{cog}}$ with SAFE-GRPO (No.6) yields the best social compliance across all settings, confirming that cognitive activations serve as essential high-level guidance for aligning RL policy improvement with human social expectations.
\item \textbf{Trade-offs in Path Efficiency (No.6 vs. No.4)}: We observe a minor drop in SPL (79.4→77.4) after RL. This reflects a fundamental trade-off:
socially compliant behaviors (e.g., following walkways, avoiding restricted regions) are often less geometrically direct. SAFE-GRPO prioritizes human-like behaviors over shortest-path efficiency, aligning with real-world navigation norms.
\end{itemize}

\vspace{-1.5mm}

\section{Conclusion}
\label{sec:conclusion}
\vspace{-1.5mm}
In this work, we presented \textbf{SocialNav}, a hierarchical foundation model capable of socially aware embodied navigation. While SocialNav provides a scalable and generalizable framework for socially intelligent navigation, several directions remain for future exploration. First, our reinforcement learning paradigm could be extended beyond semantic traversability to capture a broader range of context-dependent human conventions. Second, the current reward formulation relies on hand-crafted rules; integrating vision-language models to deliver richer, more adaptive reward signals represents a promising step toward stronger human alignment. We believe this work marks an important milestone toward embodied agents that can navigate complex, dynamic social environments with genuine social awareness.    

\clearpage

{
    \small
    \bibliographystyle{ieeenat_fullname}
    \bibliography{main}
}

\clearpage
\setcounter{section}{0}
\setcounter{page}{1}
\maketitlesupplementary
\startcontents[supplementary]
\setcounter{section}{0}
\renewcommand\thesection{\Alph{section}}
\section*{Content} 
\begingroup  
    \printcontents[supplementary]{}{1}{}
\endgroup
\vspace{1cm} 
\section{Supplementary Materials for the SocNav Dataset and Benchmark}

\subsection{Construction of Trajectories in \texorpdfstring{$D_{\text{sim}}$}{Dsim}}

\label{trajs_in_dsim}

The simulated subset $D_\text{sim}$ in the SocNav Dataset is designed to provide both \emph{standard trajectories} and \emph{recovery trajectories} in structured social scenes. This section details the construction pipeline and provides schematic visualizations.

\subsubsection{Standard Trajectory Generation}

For each scene, we first construct a navigation graph over the legal road network (sidewalks, crosswalks, plazas, etc.):

\begin{enumerate}

    \item \textbf{Road network annotation.} 
    For SocCity, we derive the traversable occupancy map $M_\text{occ}$ by mapping the semantic ground annotations inherent to the 3D city assets. For other 3D scenes (SocialGS, MatterPort3D, InteriorGS), we manually annotate $M_\text{occ}$. We ensure a unified definition of traversability across all datasets (refer to Sec.\ref{app.anno} for details). Subsequently, guided by the $M_\text{occ}$ of each scene, we manually annotate the road network along the centerlines of the traversable paths.
    
    \item \textbf{Trajectory Generation.} 
    We randomly sample a start position $\bm{s}$ and a goal position $\bm{g}$ on the traversable road network, ensuring that the geodesic distance $d(\bm{s}, \bm{g})$ along the network exceeds a minimum threshold $\ell_\text{min} = 50\,\text{m}$. We then employ an A* planner to compute the shortest path, which serves as the standard trajectory $\tau^\star$. These trajectories represent ideal navigation behavior that consistently maintains a safe distance from untraversable areas.

\end{enumerate}

\subsubsection{Recovery Trajectory Generation}

While standard trajectories provide an ideal reference, training a policy exclusively on such expert data can lead to brittleness. The agent may fail to recover from even small deviations, a classic problem of covariate shift in imitation learning. To address this and equip the agent with robust recovery capabilities, we augment the standard expert demonstrations with a set of \emph{recovery trajectories}. These are designed to simulate scenarios where the agent starts in a sub-optimal or dangerous state and must execute a corrective maneuver to rejoin the ideal path.

The generation process for these recovery trajectories is as follows:

\begin{enumerate}
    \item \textbf{Define Convergence Point.} For each standard trajectory $\tau^\star$ starting at $\bm{s}$, we first identify a convergence point $\bm{p}_\text{conv}$ on $\tau^\star$, located 5 meters ahead of the start point $\bm{s}$.

    \item \textbf{Sample Recovery Start.} We then generate a corresponding recovery start point $\bm{s}_\text{rec}$ by applying a random lateral offset (either left or right) to the standard start point $\bm{s}$. To construct a challenging recovery task, the initial orientation at $\bm{s}_\text{rec}$ is deliberately misaligned:
    \begin{itemize}
        \item If $\bm{s}_\text{rec}$ is offset to the \textbf{left}, the initial heading is set to a random angle within $[-90^\circ, -45^\circ]$ relative to the forward direction of the standard path.
        \item If $\bm{s}_\text{rec}$ is offset to the \textbf{right}, the initial heading is set to an angle within $[45^\circ, 90^\circ]$.
    \end{itemize}
    This forces the agent to start by facing away from the correct path, requiring an immediate and decisive turn.

    \item \textbf{Generate Recovery Trajectory.} A dense recovery path $\tau^\text{rec}$ is generated by linearly interpolating between the recovery start state $\bm{s}_\text{rec}$ and the convergence point $\bm{p}_\text{conv}$ with a fixed spatial step of approximately $5\,\text{cm}$. To simulate natural human micro-corrections and avoid trivial straight-line paths, we perturb each interpolated point $\bm{q}_t$ with small, zero-mean Gaussian noise:
    \begin{equation}
        \bm{q}'_t = \bm{q}_t + \bm{\epsilon}_t,\quad \bm{\epsilon}_t \sim \mathcal{N}(\bm{0}, (0.01\,\text{m})^2\mathbf{I}).
    \end{equation}
    The final perturbed path $\tau^{\text{rec}'}$ is formed by concatenating $\tau^\text{rec}$ with the path from $\bm{p}_\text{conv}$ to $\bm{g}$. It is kept only if all its points remain in cells of $\mathcal{M}_\text{occ}$  with sufficient clearance from obstacles.
    
\end{enumerate}

These systematically constructed recovery trajectories mimic plausible failure scenarios, such as starting from the wrong orientation or position, and provide explicit supervision on how to execute safe and efficient corrective actions. This enriches the training data far more effectively than simple random noise, significantly improving the policy's robustness in challenging situations.

\subsubsection{Visualization of Standard and Recovery Trajectories}

\begin{figure}[t]
    \centering
    \includegraphics[width=\linewidth]{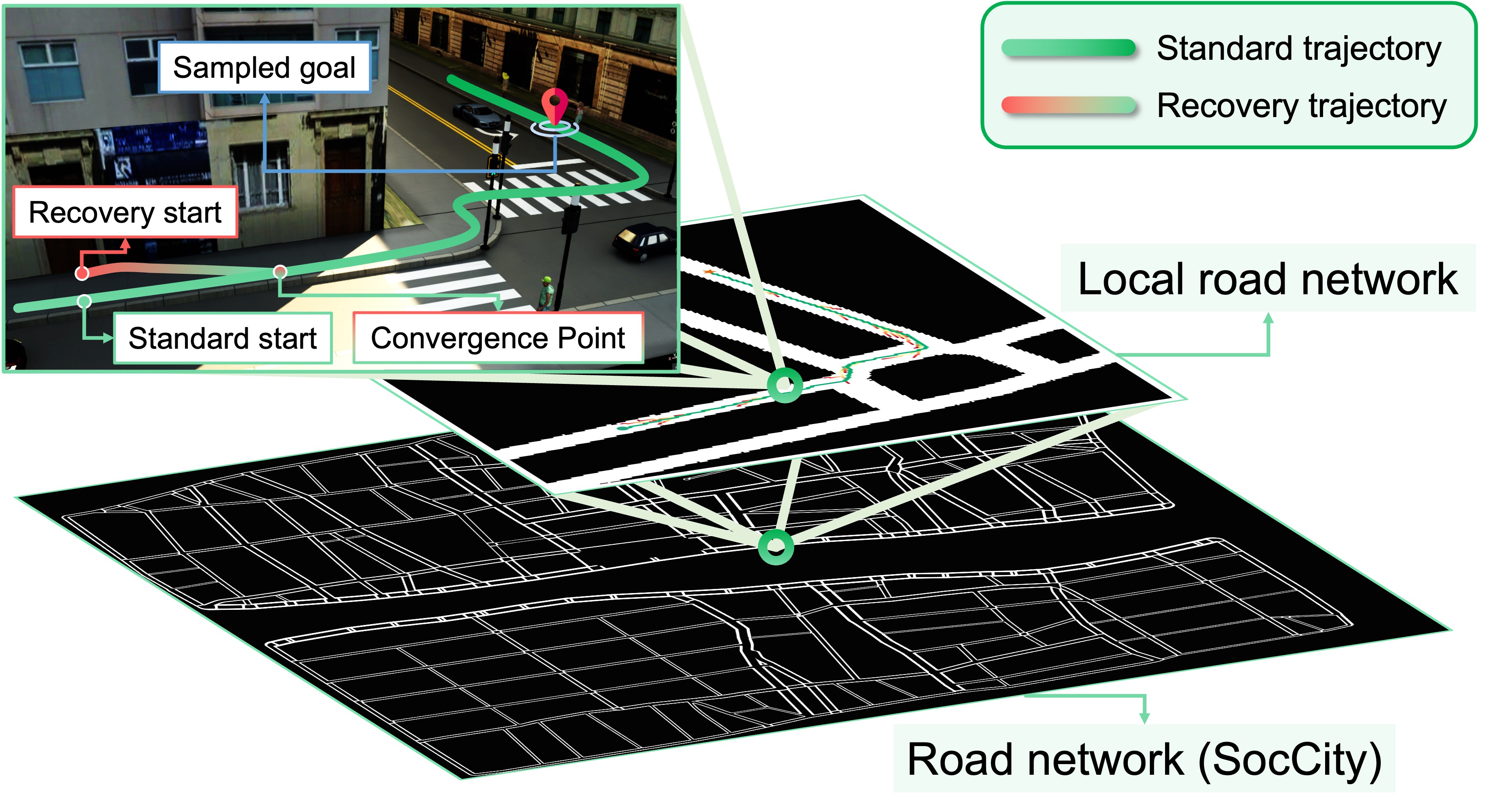}
    \caption{\textbf{Standard and recovery trajectories in $D_\text{sim}$.} Visual examples from SocCity scenes. Green curves denote standard expert trajectories obtained by A* planning on the navigation graph, while the other curves depict locally sampled recovery trajectories originating from intermediate points. The background shows the semantic occupancy map $\mathcal{M}_\text{occ}$ with walkable regions (white) and non-traversable regions (black).}
    \label{fig:supp_dsim_trajs}
\end{figure}

Figure~\ref{fig:supp_dsim_trajs} illustrates typical standard and recovery trajectories on $\mathcal{M}_\text{occ}$ , highlighting the diversity introduced by local recovery paths.

\subsection{Annotation of Socially Traversable Regions}
\label{sec:supp_traversable_regions}

Socially traversable regions provide the supervision signal for training the SocialNav Brain to predict socially compliant polygons. We describe the annotation pipeline and the corresponding guidelines.

\subsubsection{Annotation Pipeline}

\begin{enumerate}
    \item \textbf{Internet-scale data collection.} We gather first-person Internet videos and images depicting pedestrians moving in outdoor scenes such as streets, campuses, and parks.
    \item \textbf{Automatic filtering.} A vision-language model is used to filter out frames with unsuitable viewpoints (e.g., too close to walls, heavily occluded, or dominated by sky/ground) and to discard low-quality images.
    \item \textbf{Manual polygon annotation.} Human annotators draw coarse polygons on the remaining frames to delineate socially traversable regions. We create one or more polygons to cover all regions that is legally and socially allowed to walk.
\end{enumerate}

\begin{figure*}[htbp]
    \centering
    \includegraphics[width=\textwidth]{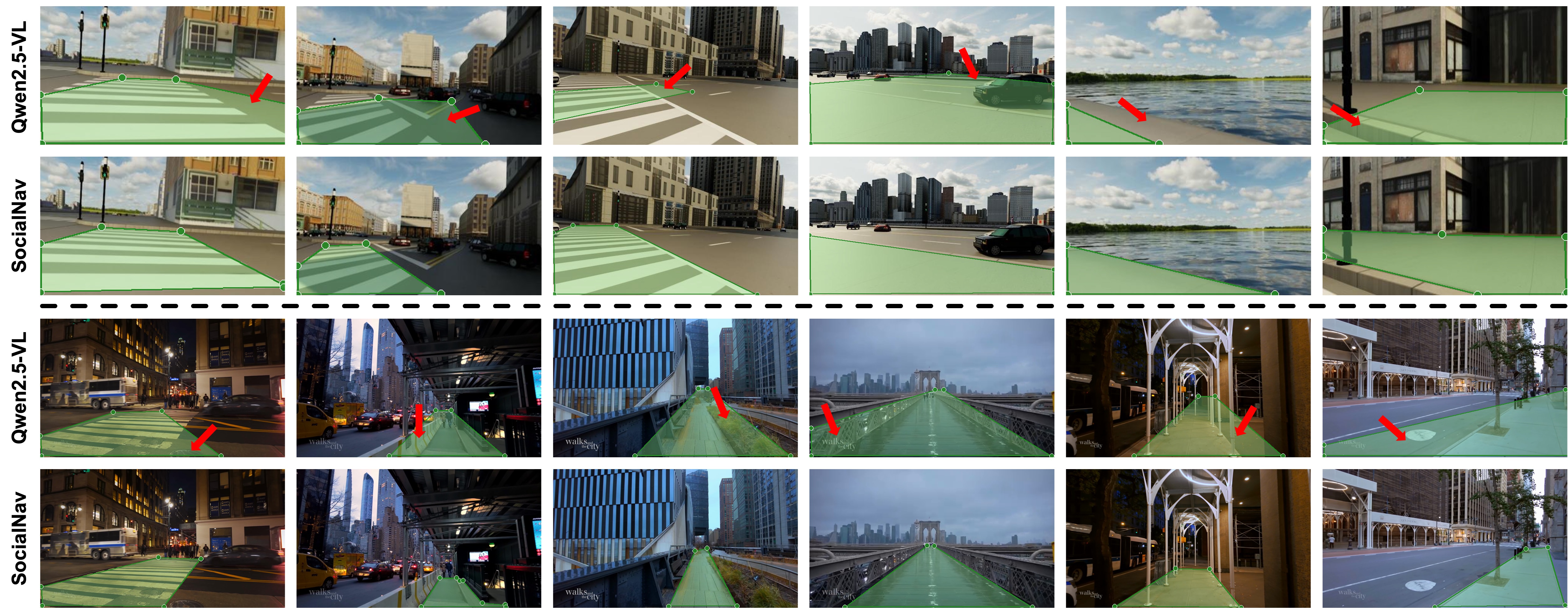}
    \caption{
\textbf{Predicted socially traversable regions on unseen scenes.}
Green polygons denote predicted socially traversable regions, and red arrows highlight areas incorrectly classified as traversable.
SocialNav yields more semantically aligned polygons in both domains.
}
    \label{fig:supp_traversable_qual}
\end{figure*}

\subsubsection{Annotation Guidelines}
\label{app.anno}

\begin{itemize}
    \item \textbf{Socially Traversable Regions.} These are defined as outdoor areas where pedestrians are permitted to walk by both legal regulations and common social norms. Annotators were instructed to label surfaces such as sidewalks, pedestrian-only streets, marked crosswalks, public plazas, and accessible outdoor staircases. For indoor scenes, all non-obstacle areas
    are considered traversable.

    \item \textbf{Non-Traversable Regions.} This category encompasses all areas that are unsafe, illegal, or socially unacceptable for pedestrian traffic. Key examples include motor vehicle lanes, bike lanes, bus lanes, green belts, ornamental lawns, flower beds, water bodies, etc.

    \item \textbf{Annotation Protocol and Polygon Standards.} Annotators draw coarse, low-vertex polygons to cover the full extent of the walkable surface; pixel-perfect alignment with curbs or markings is not required. Annotation focuses on the walkable region directly connected to the camera's viewpoint. A single polygon can span multiple connected surfaces (e.g., a sidewalk leading into a plaza). Isolated regions that cannot be reached without crossing non-traversable areas are ignored, except for pedestrian safety islands that are visibly reachable via crosswalks.
\end{itemize}

\subsubsection{Qualitative Analysis of Social Traversability Prediction}
\label{sec:supp_traversable_qual}

Figure~\ref{fig:supp_traversable_qual} presents qualitative comparisons of predicted socially traversable regions across six unseen test scenes. Each column corresponds to a different scene, while the four rows show: (1) Qwen2.5-VL predictions in SocCity, (2) SocialNav predictions in SocCity, (3) Qwen2.5-VL predictions in the real world, and (4) SocialNav predictions in the real world. Green polygons denote model-predicted socially traversable regions.

Across both simulated and real scenes, the base Qwen2.5-VL-3B model often produces polygons that are coarse, spatially misaligned, or leak into socially invalid areas such as curb edges, grass, or vehicle lanes. In contrast, the SocialNav Brain, trained with our large-scale traversability annotations, provides predictions that are more structured and socially consistent: polygons more tightly follow sidewalks, pedestrian paths, and crosswalks, and avoid visually similar but non-walkable regions. The improvements hold across all six diverse unseen scenes, indicating that fine-tuning not only enhances accuracy but also significantly strengthens cross-domain generalization.

\begin{figure*}[htbp]
\fbox{%

\begin{minipage}{\dimexpr\textwidth-2\fboxsep-2\fboxrule\relax}
\ttfamily\footnotesize
Role and task description:\\
You are now the \emph{Thinking Module} of a professional navigation AI system. Your single core task is to generate, for every movement of a quadruped robot dog, a comprehensive, in-depth, and logically rigorous \emph{Chain of Thought (CoT)}, and then output a clear movement decision. This CoT must integrate all available environmental information and explain why the final decision is the optimal choice under the current situation.\\[0.5\baselineskip]
Workflow (you \emph{must} strictly follow these three steps, without omission or reordering):\\
1.~Comprehensive analysis: parse all input information, including the robot's own state, the goal position, and environmental perception.\\
2.~CoT generation: based on your analysis, build a structured, logically sound reasoning process and evaluate all plausible alternatives.\\
3.~Final decision: select one and only one action from the candidate action space, and output it in the specified format.\\[0.5\baselineskip]
Input information:\\
1.~Scene information:\\
a.~Format: \texttt{[[history\_coord\_1], [history\_coord\_2], [history\_coord\_3], [history\_coord\_4], [current\_coord], [goal\_coord]]} together with the corresponding observation images in the same order.\\
b.~Interpretation: this sequence represents the robot dog's past and current coordinates, followed by the target coordinate.\\
c.~Coordinate system: the positive $Y$ direction corresponds to forward, and the positive $X$ direction corresponds to right.\\[0.25\baselineskip]
2.~Candidate action space:\\
\hspace*{1em}-- Go Straight\\
\hspace*{1em}-- Stay Still\\
\hspace*{1em}-- Move Backward\\
\hspace*{1em}-- Move Forward-Left\\
\hspace*{1em}-- Move Forward-Right\\
\hspace*{1em}-- Move Left\\
\hspace*{1em}-- Move Right\\[0.5\baselineskip]
Output requirements:\\
You must generate a \emph{single}, continuous block of text (no meaningless spaces, blank lines, or extra line breaks). Absolutely no additional explanations, headings, greetings, or apologies are allowed. You must strictly follow the format below:\\
\texttt{[CoT]...\;[\textbackslash CoT][Decision]...\;[\textbackslash Decision]}
\end{minipage}
}
\caption{\textbf{Navigation Chain-of-Thought Prompt Template.} The full prompt used to construct the Cognitive Activation Dataset.}
\label{fig:supp_cot_prompt}
\end{figure*}

\vspace{0.5cm}

\begin{figure*}[htbp]
    \centering
    \includegraphics[width=\textwidth]{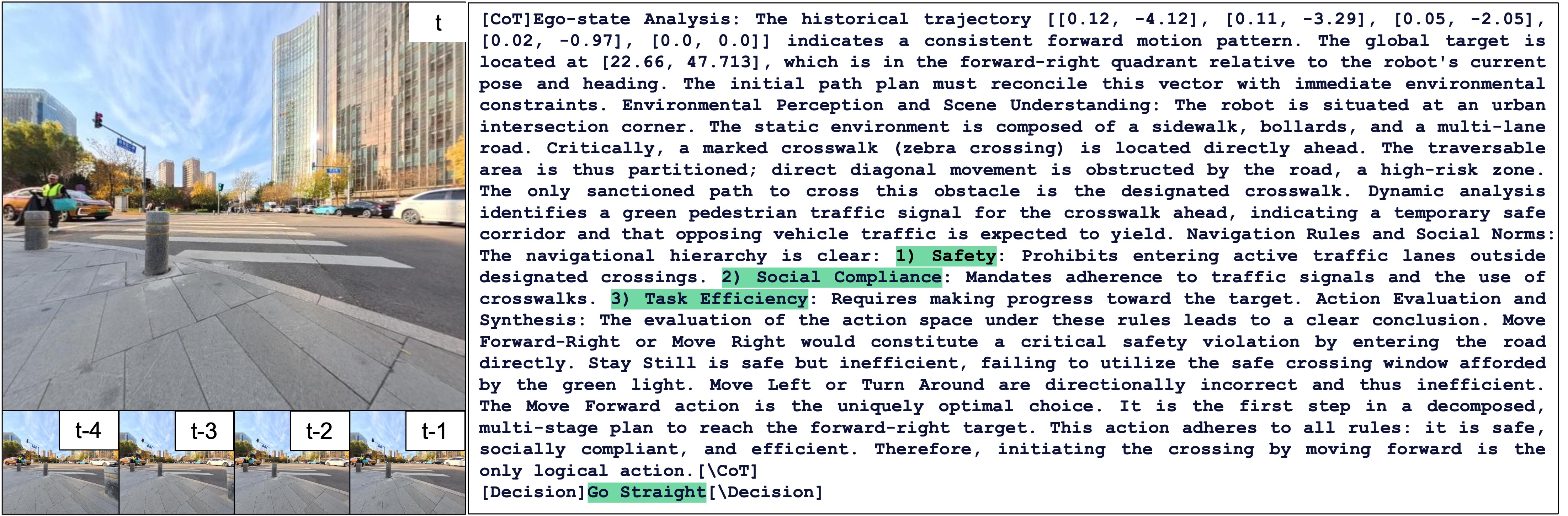}
    \caption{\textbf{Example of a navigation chain-of-thought (CoT) in an unseen crossing scenario.} Given the historical observations (left) and a goal in the forward-right direction, the Brain Module generates a structured CoT (right). The CoT demonstrates hierarchical decision-making, prioritizing Safety (no jaywalking) and Social Compliance (using the crosswalk) over direct path efficiency.}
    \label{fig:supp_cot}
\end{figure*}

\vspace{0.5cm}

\subsection{Navigation Chain-of-Thought Construction}
\label{sec:supp_nav_cot_construction}

To construct the Cognitive Activation Dataset (CAD) for navigation, we elicit chain-of-thought (CoT) style explanations using a instruction prompt. The goal is to obtain, for each navigation step, (i) a structured reasoning trace that explains the agent's decision in terms of scene layout, social norms, and future consequences, and (ii) a final discrete action selected from a predefined action space. This subsection details the prompt design and the corresponding input--output specification.

\subsubsection{Prompt Design}

\paragraph{Task and Role Description.}
We explicitly cast the brain model as the \emph{Thinking Module} of a professional navigation system. Its responsibility is to produce a logically coherent CoT and a final movement decision.

Then the prompt enforces a strict three-stage reasoning protocol:
\begin{enumerate}
    \item \textbf{Global situation analysis}: jointly parse all input information, including the robot state, the target location, and visual observations.
    \item \textbf{Chain-of-thought generation}: A \textbf{CoT segment}, enclosed between \texttt{[CoT]} and \texttt{[\textbackslash CoT]}, containing the full reasoning process which includes structured, logically tight reasoning process that evaluates the consequences of different candidate actions under social and geometric constraints.
    \item \textbf{Final decision}: A \textbf{Decision segment}, enclosed between \texttt{[Decision]} and \texttt{[\textbackslash Decision]}, containing exactly one chosen action from the predefined action space.
\end{enumerate}

\paragraph{Prompt Template.}

Figure~\ref{fig:supp_cot_prompt} presents the complete English-language prompt template used to generate the navigation CoTs for our CAD. The resulting CoT and Decision segments are stored alongside the visual observations and trajectory states, forming rich supervision for the Brain Module to acquire socially aware navigation reasoning.

\subsubsection{Qualitative Example of Navigation CoT}

Figure~\ref{fig:supp_cot} provides a qualitative example of the chain-of-thought (CoT) generated by our Brain Module for a navigation decision in an unseen urban intersection. The CoT showcases the model's ability to perform complex, multi-stage reasoning by integrating its state, goal, and rich perceptual understanding.

In this example, the model correctly identifies a conflict between the long-term goal vector (to the forward-right) and the immediate safety and social constraints. Instead of pursuing a direct path, the model decomposes the problem, recognizing that the sanctioned crosswalk directly ahead, aided by a green pedestrian signal, is the necessary intermediate step. It explicitly evaluates actions against a learned hierarchy of navigational rules—1) Safety, 2) Social Compliance, and 3) Task Efficiency—to invalidate unsafe actions (e.g., \texttt{Move Forward-Right}) and inefficient ones (e.g., \texttt{Stay Still}). The final decision to \texttt{Go Straight} is thus not a simple reactive choice but the output of a deliberate plan to safely and compliantly progress toward the goal.

\subsection{Details of the SocNav Benchmark}
The SocNav Benchmark provides a closed-loop evaluation platform for jointly assessing navigation performance and social compliance.

\subsubsection{Metric Definitions} 
\paragraph{Standard point-goal Navigation metrics.}
\begin{itemize}
    \item \textbf{MAOE.}
    Following CityWalker~\cite{liu2025citywalker}, we compute the Average Orientation Error (AOE) at each prediction step $k$ as the mean angular difference between the predicted action $\hat{a}_{i_k}$ and the ground-truth action $a_{i_k}$ over all samples:
\begin{equation}
    \text{AOE}(k) = \frac{1}{n}\sum_{i=1}^n \theta_{i_k},
    \quad
    \theta_{i_k} = \arccos\left(
        \frac{\langle \hat{a}_{i_k}, a_{i_k}\rangle}
        {\|\hat{a}_{i_k}\|_2\,\|a_{i_k}\|_2}
    \right).
\end{equation}
The Maximum Average Orientation Error (MAOE) aggregates over predicted horizons:
\begin{equation}
    \text{MAOE} = \frac{1}{n}\sum_{i=1}^n \max_k \theta_{i_k},
\end{equation}
which emphasizes the worst-case orientation mismatch along future steps.
    \item \textbf{Success Rate (SR).} The fraction of episodes in which the agent reaches within $3\,\text{m}$ of the target with fewer than three collisions.
    \item \textbf{Route Completion (RC).} The ratio between the geodesic distance from start to the final position and the geodesic distance from start to goal.
    \item \textbf{Success weighted by Path Length (SPL).} The SPL metric, as in prior navigation work, jointly captures success and path efficiency.
\end{itemize}

\paragraph{Social compliance metrics: DCR and TCR.}
The Distance Compliance Rate (DCR) measures the proportion of distance spent in socially traversable regions:
\begin{equation}
\text{DCR} =
\begin{cases}
    \dfrac{d_\text{compliant}}{d_\text{actual}}, & \text{if } s = 1,\\[4pt]
    0, & \text{otherwise},
\end{cases}
\end{equation}
where $s \in \{0,1\}$ indicates success, $d_\text{compliant}$ is the distance traveled in regions labeled as socially traversable, and $d_\text{actual}$ is the total traveled distance. 

The Time Compliance Rate (TCR) is defined analogously by replacing distance with duration spent in compliant regions. Both metrics are computed only for successful episodes and are averaged over all tasks.

\subsubsection{Pedestrian Placement and Behavior Model}
\label{sec:supp_agents}

To simulate realistic social environments, we populate the evaluated scenes with dynamic pedestrians:
\begin{itemize}
    \item \textbf{Placement and Density.} To ensure realistic positioning, pedestrians are spawned within a specific margin, between 50 cm and 100 cm from the inner boundary of the socially traversable regions. We maintain a pedestrian density of not exceeding 6 individuals per 100 meters of walkable path.
    \item \textbf{Behavior model.} The behavior of each pedestrian is governed by a skeletal animation system, which supports both walking and running actions. Each pedestrian selects a random goal on the road network and moves along the shortest path at a speed sampled from a truncated normal distribution (mean $1.0\,\text{m/s}$, standard deviation $0.2\,\text{m/s}$).
    
    \item \textbf{Interaction with the robot.} Pedestrians do not explicitly cooperate with the robot; they follow their own routes. Consequently, the robot must proactively adapt its trajectory to maintain social distances and avoid intruding into non-traversable regions.
\end{itemize}


\section{Supplementary Materials for the SocialNav Foundation Model}
The \textbf{SocialNav} foundation model adopts the hierarchical Brain--Action design described in the main paper. Here we summarize key architectural details.

\subsection{Model Architecture and Parameters}
\label{sec:model_arch}

Table~\ref{tab:model_params} summarizes the architectural parameters of the \textbf{Brain Module} (Qwen2.5-VL-3B) and the \textbf{Action Expert} (Diffusion Transformer).  
The table includes structural details necessary for reproducibility.

\begin{table}[t]
\centering
\caption{\textbf{Model Architecture and Parameters.}}
\label{tab:model_params}
\renewcommand{\arraystretch}{1.0}
\setlength{\tabcolsep}{4pt}
\begin{tabular}{l|c}
\toprule
\multicolumn{2}{c}{\textbf{Brain Module (Qwen2.5-VL-3B)}} \\
\midrule
Layers / Hidden / MLP & 36 / 2048 / 11008 \\
Attention heads (KV) & 16 (2) \\
Activation / Norm & SiLU / RMSNorm \\
Max context & 128k (SW: 32k) \\
RoPE scaling & multi-segment (16/24/24) \\
Vocab size & 151{,}673 \\
\midrule
\multicolumn{2}{c}{\textbf{Action Expert (Diffusion Transformer)}} \\
\midrule
Action dim / Chunk size & 2 / 5 \\
Flow-matching steps & 5 \\
\bottomrule
\end{tabular}
\end{table}

\subsection{Training Details}
\label{sec:training_details}

This section provides concise supplementary training information for the three-stage pipeline introduced in the main paper.  
Table~\ref{tab:three_stage_hparams} summarizes both the trainable components and the datasets used in each stage, while some shared optimization settings remain unchanged across the entire pipeline.

\begin{table}[t]
\footnotesize
\centering
\caption{\textbf{Trainable Components and Datasets Across Stage~1--3.}  
Checkmarks indicate modules updated during each stage.}
\label{tab:three_stage_hparams}
\resizebox{\linewidth}{!}{%
\renewcommand{\arraystretch}{1.10}
\begin{tabular}{l|c|c|c|c|c}
\toprule

\multirow{3}{*}{\textbf{Stage}} 
& \multirow{3}{*}{\textbf{Datasets}} & \multicolumn{4}{c}{\textbf{Trainable Modules}} \\
& 
& \makecell{\textbf{VLM}\\\textbf{Brain}} 
& \makecell{\textbf{Vision}\\\textbf{Encoder}} 
& \makecell{\textbf{Waypoint}\\\textbf{Encoder}} 
& \makecell{\textbf{Action}\\\textbf{Expert}} \\
\midrule
\textbf{Pre-training} 
& $D_\text{video}$, $D_\text{sim}$, $D_\text{cog}$ 
& \checkmark & \checkmark & \checkmark & \checkmark \\

\textbf{Fine-tuning} 
& $D_\text{real}$ 
& \texttimes & \texttimes & \texttimes & \checkmark \\

\textbf{SAFE-GRPO} 
& $D_\text{sim}$ (SocCity)
& \texttimes & \texttimes & \texttimes & \checkmark \\
\bottomrule
\end{tabular}
}
\end{table}

Hyperparameter settings are summarized in Table~\ref{tab:shared_hparams}.

\begin{table}[t]
\centering
\caption{\textbf{Hyperparameter Settings.}}
\label{tab:shared_hparams}
\renewcommand{\arraystretch}{1.12}
\setlength{\tabcolsep}{7pt}
\begin{tabular}{l|c}
\toprule
\textbf{Settings} & \textbf{Value} \\
\midrule
Optimizer & AdamW \\
Adam betas $(\beta_1,\beta_2)$ & $(0.9, 0.95)$ \\
Weight decay & $0.1$ \\
LR scheduler & Cosine decay \\
Precision & BF16 \\
Flash Attention 2 & Enabled \\
Gradient accumulation & $1$ \\
Gradient checkpointing & Enabled \\
\bottomrule
\end{tabular}
\end{table}

\medskip

\begin{table*}[htbp]
\centering
\caption{\textbf{Ablation on SAFE-GRPO reward components on the SocNav Benchmark.} We deactivate each reward term in turn by setting its weight to zero. Checkmarks indicate that the reward is used.}
\label{tab:supp_reward}
\begin{tabular}{l|cccc|ccccc}
\toprule
\multirow{2}{*}{\textbf{Variant}} 
& \multicolumn{4}{c|}{\textbf{Reward terms}} 
& \multicolumn{5}{c}{\textbf{SocNav Benchmark}} \\
\cmidrule(lr){2-5}\cmidrule(lr){6-10}
& $\mathcal{R}_\text{social}$ 
& $\mathcal{R}_\text{expert}$ 
& $\mathcal{R}_\text{smooth}$ 
& $\mathcal{R}_\text{effciency}$ 
& SR$\uparrow$ & RC$\uparrow$ & SPL$\uparrow$ & DCR$\uparrow$ & TCR$\uparrow$ \\
\midrule
w/o $\mathcal{R}_\text{social}$ 
& --          & \checkmark & \checkmark & \checkmark 
& 84.7 & 90.3 & \textbf{78.5} & 61.4 & 62.1 \\

Full SAFE-GRPO 
& \checkmark & \checkmark & \checkmark & \checkmark 
& \textbf{86.1} & \textbf{91.2} & 77.4 & \textbf{82.5} & \textbf{82.9} \\
\bottomrule
\end{tabular}
\end{table*}

\begin{table*}[htbp]
    \centering
    \caption{\textbf{Open-Loop Evaluation on CityWalker Benchmark}~\cite{liu2025citywalker}. We evaluate MAOE metric in each critical scenario for all methods. Percentages under scenarios indicate their data proportions. The "Mean" column shows scenario means averaged over six scenarios; "All" shows sample means over all data samples.}\label{tab:supply_benchmark}
    \begin{tabular}{l|cccccccc}
    \toprule

        \multirow{2}{*}{\textbf{Method}} & \multirow{2}{*}{\textbf{Mean}} & \textbf{Turn} & \textbf{Crossing} & \textbf{Detour} & \textbf{Proximity} & \textbf{Crowd} & \textbf{Other} & \textbf{All} \\ 

        & & 8\% & 12\% & 12\% & 6\% & 7\% & 55\% & 100\% \\ 
         \midrule
                GNM*~\cite{shah2023gnm}
         & 15.2 & 29.5 & 13.6 & 11.9 & 13.6 & 12.3 & 10.4 & 11.5\\

        ViNT*~\cite{shah2023vint}
         & 15.8 & 30.0 & 14.6 & 12.3 & 14.0 & 12.7 & 11.0 & 12.2\\

        NoMaD*~\cite{sridhar2024nomad}
         & 17.8 & 33.4 & 16.5 & 14.8 & 16.2 & 13.6 & 12.2 & 12.3\\

        CityWalker*~\cite{liu2025citywalker}
         & 14.2 & 25.0 & 12.8 & 12.5 & 13.2 & 11.5 & 10.0 & 11.2\\
         \midrule
        \rowcolor{lavender}
        SocialNav*
         &  11.7 & 21.5 &  10.9  &  10.4 &  9.8 &  8.9 &  8.7 &  9.4\\ 
        \rowcolor{lavender}
        \textbf{SocialNav (Full)}
         &  \textbf{10.2} & \textbf{20.1} &  \textbf{8.8}  &  \textbf{8.4} &  \textbf{8.9} &  \textbf{7.6} &  \textbf{7.2} &  \textbf{7.8}\\ 
    \bottomrule
    \end{tabular}
\end{table*}

\subsection{SAFE-GRPO Reward Design}
\label{reward_detail}

\noindent\textbf{Social Compliance Reward $\mathcal{R}_\text{social}$}: 
This is the primary incentive for respecting both physical safety and social norms. 
From $\mathcal{M}_\text{occ} \in \{0,1\}^{H \times W}$, we compute a Distance Transform (DT) map $D(\bm{x})$, which assigns each traversable location $\bm{x}$ its Euclidean distance to the nearest non-traversable cell. This DT map encodes both collision avoidance and social distancing principles—higher values indicate safer, more normatively acceptable regions.

Let $\{\bm{x}_t\}_{t=1}^T$ be the predicted trajectory in world coordinates, and let $\bar{d}_{\text{pred}} = \frac{1}{T} \sum_{t=1}^T D(\bm{x}_t) $ denotes the average obstacle-free clearance along the path. Similarly, we compute $\bar{d}_{\text{gt}}$ for the expert trajectory.
The social reward is then formulated as:
\begin{equation}
    \mathcal{R}_{\text{social}} = \beta \cdot \sigma\left( \frac{\bar{d}_{\text{pred}} - \bar{d}_{\text{gt}}}{\alpha} \right),
    \label{eq:reward_social}
\end{equation}
where $\sigma(\cdot)$ denotes the sigmoid function, and hyperparameters $\alpha=0.5$, $\beta=2.0$ control sensitivity and scaling. This formulation rewards trajectories that maintain comparable or greater clearance than the expert.

\noindent\textbf{Expert Trajectory Similarity Reward ($\mathcal{R}_{\text{expert}}$):} 
We measure similarity in both spatial proximity and directional consistency. Given predicted trajectory $\bm{p}$ and expert $\bm{g}$ in world coordinates:

\begin{equation}
    \mathcal{R}_{\text{expert}} = w_d \cdot r_{\text{dist}} + w_\theta \cdot r_{\text{dir}},
    \label{eq:reward_expert}
\end{equation}
where:
\begin{align}
    r_{\text{dist}} &= \exp\left(-\frac{1}{T}\sum_{t=1}^T \|\bm{p}_t - \bm{g}_t\| \Big/ \tau_d \right),  \\
    r_{\text{dir}}  &= \frac{1}{2} \left( \cos(\Delta\theta_{\text{avg}}) + 1 \right) \in [0,1],
\end{align}
with $w_d = 0.7$, $w_\theta = 0.3$, $\tau_d = 1.0\,\text{m}$. Here, $\Delta\theta_{\text{avg}}$ is the average angular difference between consecutive displacement vectors. 

\noindent\textbf{Trajectory Smoothness Reward ($\mathcal{R}_{\text{smooth}}$):} 
We encourage consistent step lengths by penalizing high variance in inter-step distances:
\begin{equation}
    \mathcal{R}_{\text{smooth}} = \exp\left( -\frac{\mathrm{std}(\{\|\Delta \mathbf{x}_t\|\}_{t=2}^T)}{\alpha_s} \right), \quad 
    \label{eq:reward_smooth}
\end{equation}
where $\alpha_s = 0.8$, and $\mathrm{std}(\cdot)$ computes the standard deviation of step magnitudes. A lower variance yields higher reward, promoting natural gait-like movement.

\noindent\textbf{Path Efficiency Reward ($\mathcal{R}_{\text{efficiency}}$):} 
To encourage forward progress without excessive detours, we compare the agent’s net advancement to that of the expert:
\begin{equation}
\resizebox{0.46\textwidth}{!}{%
    $ \displaystyle
        \mathcal{R}_{\text{efficiency}} = \beta_l \cdot \sigma\left( \frac{\|\bm{x}_T - \bm{x}_0\|_2 - \|\bm{x}^{\text{gt}}_T - \bm{x}^{\text{gt}}_0\|_2}{\alpha_l} \right), \quad
    $
    \label{eq:reward_length}
}
\end{equation}

where $\alpha_l = 5.0$, $\beta_l = 2.0$. By combining these reward components, our design ensures that the agent learns to navigate not only effectively, but also in a manner that is predictable, respectful, and aligned with human expectations within shared environments.

\subsection{Ablation on SAFE-GRPO Reward Funtions}
\label{sec:supp_reward_ablation}

We ablate the $\mathcal{R}_\text{social}$ defined in Eq.~(4) of the main paper. For the variant, we set the weight of the $\mathcal{R}_\text{social}$ to zero while keeping all other training configurations unchanged.

The results in Table~\ref{tab:supp_reward} show that $\mathcal{R}_\text{social}$ is crucial for high DCR and TCR; without it, the agent tends to take shorter but socially risky shortcuts.

\begin{figure*}[htbp]
    \centering
    \includegraphics[width=0.95\textwidth]{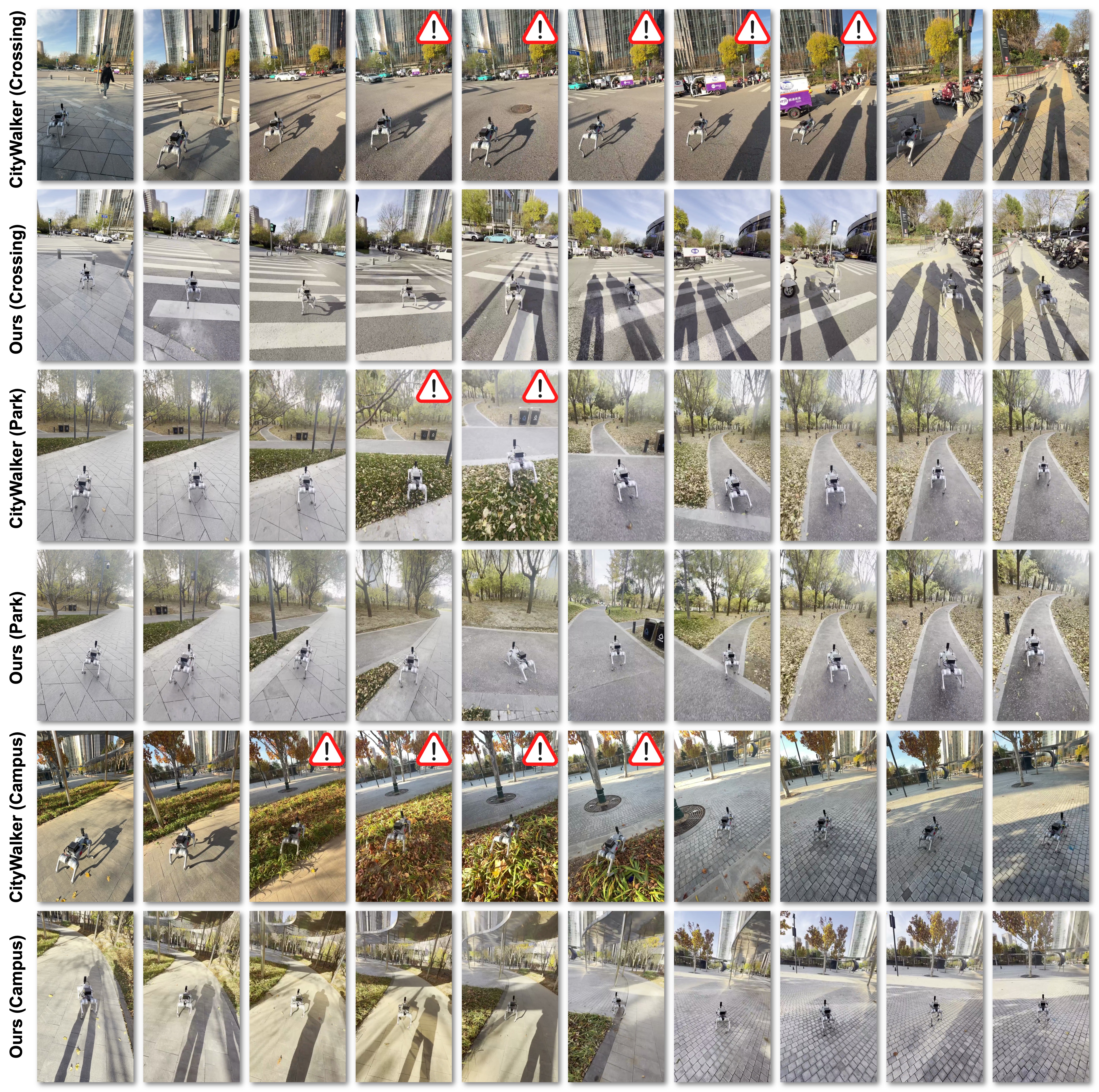}
    \caption{\textbf{Real-world deployment visualizations.} Third-person views of the Unitree Go2 robot navigating in street, park, and campus environments. SocialNav successfully follows sidewalks, avoids stepping onto lawns or driveways, respects pedestrian flows.}
    \label{fig:supp_real_robot}
\end{figure*}

\section{Additional Experimental Results}

\subsection{Extended Open-Loop Results on the CityWalker Benchmark}
\label{sec:supp_citywalker_openloop}

We provide an extended comparison on the CityWalker open-loop benchmark.
The model variants marked with asterisk (*) denotes the models that were exclusively trained on the $D_\text{real}$ dataset.

Compared to the official CityWalker model, the retrained CityWalker* shows consistent improvements across all scenarios, confirming that our $D_\text{real}$ provides a more comprehensive navigation motion priors. Similarly, GNM*, ViNT*, and NoMaD* benefit from retraining.

SocialNav (Full) achieves the lowest MAOE in every scenario and in the overall mean. The largest relative gains are observed in the \emph{Turn} and \emph{Crossing} categories, where understanding of social layout and high-level semantics is particularly important. This trend supports our claim that integrating the Brain Module with the flow-based Action Expert leads to trajectories that more closely match human-operated behaviors.

\subsection{Additional Real-World Deployment Visualizations}
\label{sec:supp_realworld_more}

To further illustrate the real-world performance of SocialNav, we visualize third-person view from the Unitree Go2 deployments described in Table~\ref{tab:real_world_results}.

The visualizations highlight that SocialNav can be executed in real-time on a cloud server with an NVIDIA A10 GPU, maintaining over $5$ Hz control frequency while preferring socially acceptable walkways even when shorter but socially inappropriate shortcuts exist.

\end{document}